\documentclass[letterpaper, 10 pt, conference]{ieeeconf}  

\IEEEoverridecommandlockouts                              

\usepackage{balance}
\usepackage{mathtools}
\usepackage{hyperref}
\usepackage{xcolor}
\usepackage[font=footnotesize,hypcap=false]{caption}
\usepackage{siunitx}
\usepackage{balance}
\usepackage{multirow}
\usepackage{amsfonts}

\usepackage{caption}
\usepackage{subcaption}

\title{\LARGE \bf
  Robust Humanoid Walking on Compliant and Uneven Terrain \\
  with Deep Reinforcement Learning
}

\author{Rohan P. Singh$^{1,2}$, Mitsuharu Morisawa$^{1}$, Mehdi Benallegue$^{1}$, \\
  Zhaoming Xie$^{3}$, Fumio Kanehiro$^{1,2}$
\\\tt\small{Email: rohan-singh@aist.go.jp}
\thanks{
$^{1}$
CNRS-AIST JRL (Joint Robotics Laboratory) IRL,
National Institute of Advanced Industrial Science and Technology (AIST),
Japan.}
\thanks{
$^{2}$
University of Tsukuba,
Ibaraki,
Japan.}
\thanks{
$^{3}$
Department of Computer Science,
Stanford University,
USA.}
}%

\begin{document}

\maketitle
\thispagestyle{empty}
\pagestyle{empty}

\renewcommand*{\thefootnote}{\fnsymbol{footnote}}

\begin{abstract}
  For the deployment of legged robots in real-world environments, it is essential to develop
  robust locomotion control methods for challenging terrains that may exhibit unexpected
  deformability and irregularity.
  In this paper, we explore the application of \textit{sim-to-real} deep reinforcement learning (RL)
  for the design of bipedal locomotion controllers for humanoid robots on compliant and uneven terrains.
  Our key contribution is to show that a simple training curriculum for exposing the RL agent
  to randomized terrains in simulation can achieve robust walking on a real
  humanoid robot using only proprioceptive feedback.
  We train an end-to-end bipedal locomotion policy using the proposed approach,
  and show extensive real-robot demonstration on the HRP-5P humanoid over several difficult terrains inside and outside
  the lab environment.

  Further, we argue that the robustness of a bipedal walking policy can be improved if the
  robot is allowed to exhibit aperiodic motion with variable stepping frequency.
  We propose a new control policy to enable modification of the observed clock signal,
  leading to adaptive gait frequencies depending on the terrain and command velocity.
  Through simulation experiments, we show the effectiveness of this policy specifically for walking
  over challenging terrains by controlling swing and stance durations.

  The code for training and evaluation is available online
  \footnote[2]{\url{https://github.com/rohanpsingh/LearningHumanoidWalking}}.
\end{abstract}

\renewcommand*{\thefootnote}{\arabic{footnote}}

\section{Introduction}

Uncertainties in terrain properties --- such as the height profile and deformability ---
present significant challenges for conventional model-based approaches to humanoid locomotion.
This is primarily due to the strict temporal and spatial assumptions placed by such approaches
on the foot trajectories and environmental contacts \cite{kajita2010biped, caron2019stair}.
When faced with an irregular or compliant (i.e. deformable) surface, these assumptions may be violated because of
premature or delayed foot landing leading to negative consequences on the control. This is especially critical for humanoid robots with their large mass and bipedal support on
bulky legs.

\begin{figure}[t]
  \center
  \includegraphics[width=0.9\linewidth]{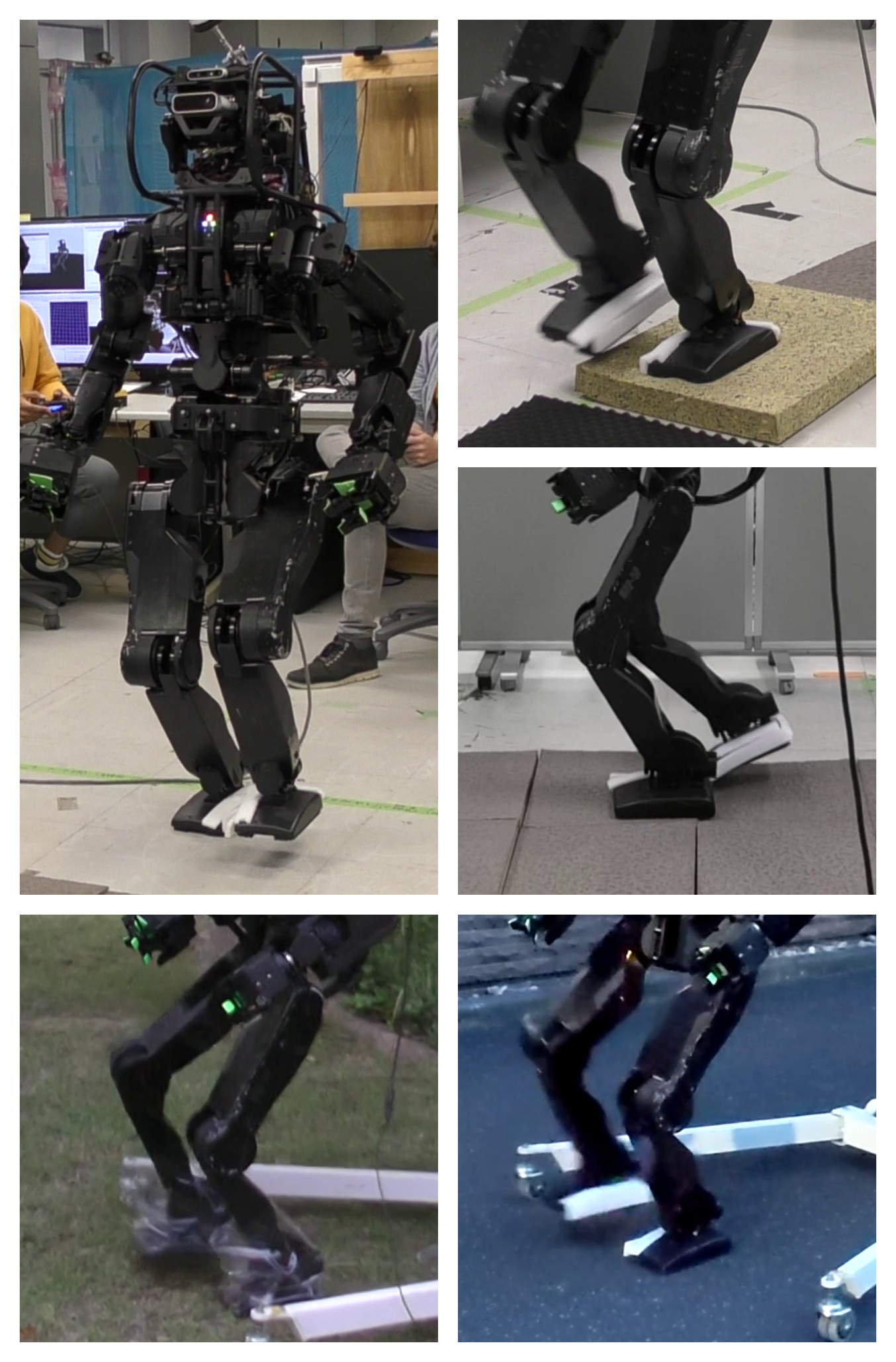}
  \caption{\textbf{HRP-5P humanoid} bipedal locomotion (clockwise) on flat rigid floor, soft cushion,
    uneven inclined blocks, paved street, and grass using learned policy. The same RL policy was used
    for all terrains without any parameter tuning in between experiments. Lifter and overhead
    crane serve as a failsafe, the ropes are slack and robot is not provided external support.
  }
  \label{figure:terrains}
\end{figure}


For real-world applications, it is important to develop a single, unified locomotion controller
that can enable the robot to walk on different terrain types without the need for gait parameter
tuning or manual specification of surface properties. Furthermore, as different terrains entail
different walking strategies and contact timings \cite{hashimoto2012realization, walas2016terrain}, the controller must
preferably be \textit{adaptive} --- to adapt the gait according to the type of terrain.

The problem becomes more complicated in the case of blind locomotion, that is, when the robot
is unable to perceive the compliance and topography of the upcoming environment using
exteroceptive sensors. In such cases, the controller needs to estimate the terrain properties from the
observed state, more specifically, from joint positions and velocities, inertial measurement
unit (IMU), clock signal, etc.

In this work, we present a locomotion controller for bipedal walking on deformable and uneven
terrains using model-free deep RL without relying on explicit identification or classification.
Instead, we let the controller learn to implicitly adapt to different terrains by exposing it
to a range of terrains during the training
phase in a simulation environment.
By using a combination of one-hot representation of the user-commanded walking mode and clock-based reward
terms \cite{siekmann2021sim}, we propose to develop a single policy that can perform multiple \textit{modes}
of walking: stepping and turning in-place, walking forward, and quit standing.

Then, we perform \textit{sim-to-real} transfer to deploy the trained policy
on the real HRP-5P humanoid robot \cite{kaneko2019humanoid} on
different types of environments, both indoors and outdoors.

Next, we propose to further enhance the robustness of the walking policy by
learning policies that can predict clock signal modulations and
achieve \textit{aperiodic} gaits. Intuitively, these modulations
further relax the constraints on the RL agent on contact timing, and consequently
alleviate the challenge of premature and delayed landing of the swing foot on unknown
terrains. We evaluate this policy
systematically in the simulation environment and demonstrate the advantage of adaptive
stepping frequency specifically for challenging terrain locomotion.


\section{Related Works}

\subsection{Conventional approaches}
Methods based on analytically-derived models of the robot locomotion dynamics
have been used for humanoid locomotion, such as the linear inverted pendulum model \cite{kajita2003biped}
and its variants and the hybrid zero dynamics \cite{westervelt2003hybrid}. Both approaches
have led to recent successful robust implementation on various kinds of terrains \cite{mesesan2019dynamic,
  reher2021inverse}.

An important work on the TORO robot \cite{mesesan2019dynamic} shows impressive locomotion
over rough terrain, grass, and a soft gym mattress. However, like several other model-based
methods, it involves manual tuning of gait parameters (timing, step length) and task
gains.

Further work has been developed specifically to tackle terrains. For instance, some works have focussed on
the optimization of gait parameters based on analysis of human experimental data \cite{
  hashimoto2012realization}, or based on terrain classification from tactile measurements
\cite{walas2016terrain}. Similarly, step adjustment based on divergent component of motion
(DCM) error on a torque-controlled humanoid has been attempted \cite{hopkins2015design}.
Other works have been limited to the simulation environment, for ground reaction force (GRF)
based control \cite{komuta2017walking}, and for simulation of sinkage phenomena on soft terrains
\cite{komizunai2010development}.

A parallel approach has been to design new feet mechanisms for better
passive and active terrain adaptation \cite{frizza2022study}.

\subsection{Data-driven approaches}
More recently, deep reinforcement learning (RL) has emerged as a radically new approach for
legged locomotion. Application of deep RL to real quadrupedal robots has quickly equalled or surpassed
model-based approaches in terms of locomotion performance in outdoor environments \cite{2020-science-blindQuadruped, miki2022learning,
  choi2023learning}. Bipedal locomotion has also been achieved with deep RL on lighter bipeds and humanoids
like Cassie \cite{2019-CORL-cassie, 2020-rss-RNNCassie} and Digit
\cite{2021-IROS-DigitRL, radosavovic2023learning}, and on smaller humanoid
platforms \cite{2021-ICRA-DeepWalk, masuda2022sim}. However, demonstrations
on life-sized humanoids like the HRP-5P \cite{singh2023learning} and TOCABI \cite{kim2023torque}
have mostly been confined to flat terrains with relatively small obstacle size.

Moreoever, while demonstrations on rigid obstacles on stiff surfaces has been shown,
learning-based approaches for compliant surfaces are more limited.

Note that the challenge posed by such terrains is especially greater for HRP-5P, compared to robots
with lighter legs \cite{2020-science-blindQuadruped, siekmann2021blind}. During the training phase
in simulation, due to the large inertia, low backdrivability, and low joint speeds, HRP-5P may find
it much more difficult to recover from an unexpected collision with an obstacle --- leading to degraded
learning. This means that exploration during training can be problematic in an environment with large
obstacles. On compliant surfaces, the surface will yield considerably more under the large total
mass of HRP-5P.

Hence, the application of
end-to-end deep RL policies for life-sized humanoids to tackle challenging terrains is a matter of
investigation.

\section{Environment}

We use model-free deep reinforcement learning to train our locomotion policies in a
simulation environment, and provide the details of the RL environment in this section.

\subsection{Observations, Actions, and Rewards.}
\textbf{Observation space.}
Similar to \cite{singh2023learning, singh2022learning}, the policy observation space includes
the robot state and the external state (refer to \autoref{table:observations}).

The robot state includes proprioceptive measurements from the encoders, on-board IMU, and motor
current sensors. 
The current measurements are scaled to joint-level torque using the gear ratio and torque constant.

To control the robot's walking speed and direction, we use a $3D$ one-hot encoding to denote the
walking mode --- $[0, 0, 1]$ for standing and $[0, 1, 0]$ for stepping in-place and $[1, 0, 0]$
for walking forward -- and a $1D$ scalar to denote the reference value for the turning speed or
the forward walking speed, depending on the mode. On the real robot, the values for the walking
mode and mode reference are obtained from the robot operator using a joystick.

A clock signal is derived from a cyclic phase variable $\phi$ as follows:

\begin{align}
\label{eq:clock}
\text{clock} = \left\{ \sin \left( \frac{2\pi\phi}{L} \right), \cos \left( \frac{2\pi\phi}{L} \right)  \right\},
\end{align}

\noindent
where $L$ is the cycle period. $\phi$ increments from 0 to 1 at each control timestep and reset to 0
after every $L$ timesteps. Clock is then used as input to the policy. We set the gait cycle duration
to 2 seconds (at $40\SI{}{\hertz}$ control frequency, $L = 2 \times 40 = 80$ timesteps).

\textbf{Action space.}
The $12D$ action space of the policy is comprised of the desired positions of the actuated joints of the
robot's legs (6 in each). The predictions from the network are added to fixed motor offsets corresponding
to a nominal posture and are then tracked using a low-gain PD controller (see \autoref{table:actions}):

\begin{align}
\label{eq:pdcontrol}
\text{tau}_{pd} = K_{p}(q_{des} - q) + K_{d}(0 - \dot{q}),
\end{align}

\noindent
where $K_{p}$ and $K_{d}$ denote the proprotional and derivative gain factors respectively. $q_{des}$ is
the policy prediction summed with the fixed motor offsets. $q$ and $\dot{q}$ denote the measured joint
position and velocity. In the simulation environment, this torque drives the direct-drive motors while
on the real robot, this is further tracked using low-level current controllers \cite{masuda2022sim, singh2023learning}.
The upperbody joints are frozen in the nominal configuration using stiff PD control. The policy is executed
at $40\SI{}{\hertz}$ while the PD loop runs at $1000\SI{}{\hertz}$, both in simulation and on real robot.

\textbf{Reward design.}
We adopt the same reward function as in \cite{singh2023learning}. The full reward function consists of
the bipedal gait terms inspired from previous works \cite{yang2020learning, siekmann2021sim}, terms
to encourage tracking of the walking mode command and reference speed, and terms for
developing realistic motion for safer \textit{sim-to-real} transfer. The terms are listed in \autoref{table:rewards}
along with their relative weights. We refer the reader to \cite{singh2023learning} and the open-source
code \footnote{\url{https://github.com/rohanpsingh/LearningHumanoidWalking}} for the precise description
of each term.

\subsection{Initialization and Termination.}  At the start of each training episode, the robot joints are
initilialized to the ``half-sitting" posture (the configuration at which the robot can remain upright
and prevent falling in the absence of external disturbances) injected with a small amount of noise
 \footnote{Initialization noise was ultimately deemed to be unnecessary during real robot experiments.} .
The phase variable $\phi$ is initilialized to $0$ and the walking mode is set to \textit{Standing}.

Early termination conditions are needed to avoid exploration in irrecoverable states. We enforce a fall
conditions, reached when the root height from the lowest point of foot-floor contact is less than $60\SI{}{\cm}$,
and a self-collision condition. An episode ends either after a fixed number of control timesteps or
when a termination condition evaluates to true.

\subsection{Target terrains.}
Our primary objective in this work is to enable the robot to walk successfully over terrains
of different compliance, small inclines, rigid obstacles, and ground unevenness. For evaluating
the real robot's locomotion capabilities in indoor experiments, we created a testbed consisting
of the following surfaces: a flat, stiff floor, a blue mattress, a block of soft cushion foam,
rigid blocks of varying inclination, height and uneveness.


For evaluations in outdoor environments, we perform demonstrations in a grass lawn (deformable
and with small obstacles) and paved street (gently inclined).

\begin{table}[t]
\caption{Policy Inputs.}
\label{table:observations}
\begin{center}
\begin{tabular}{l l c}
\hline
& \multicolumn{1}{l}{\textbf{Observation}} & \multicolumn{1}{l}{\textbf{Dimension}}\\
\hline
\multirow{5}{*}{Robot state} & root orientation (roll, pitch) & 2\\
 & root angular velocity & 3\\
 & joint positions & 12\\
 & joint velocities & 12\\
 & motor currents & 12\\
\hline
\multirow{3}{*}{External state} & Mode \textit{(Standing, Inplace, Forward)} & 3\\
 & Mode reference & 1\\
 & Clock & 2\\
\hline
\end{tabular}
\end{center}
\end{table}

\begin{table}[t]
\caption{Policy Outputs.}
\label{table:actions}
\begin{center}
\begin{tabular}{l l c c}
\hline
& \multicolumn{1}{l}{\textbf{Action}} & \multicolumn{1}{l}{\textbf{Kp}} & \multicolumn{1}{l}{\textbf{Kd}}\\
\hline
\multirow{6}{*}{Leg joints}
 & (R/L) hip yaw & 200 & 20\\
 & (R/L) hip roll & 150 & 15\\
 & (R/L) hip pitch & 200 & 20\\
 & (R/L) knee pitch & 150 & 15\\
 & (R/L) ankle pitch & 80 & 8\\
 & (R/L) ankle roll & 80 & 8\\
\hline
\end{tabular}
\end{center}
\end{table}

\begin{table}[t]
\caption{Reward Function.}
\label{table:rewards}
\begin{center}
\begin{tabular}{l l c}
\hline
& \multicolumn{1}{l}{\textbf{Term}} & \multicolumn{1}{l}{\textbf{Weight}}\\
\hline
\multirow{2}{*}{Bipedal walking (clock-based)}
 & Foot force & 0.225\\
 & Foot speed & 0.225\\
\hline
\multirow{2}{*}{Objective}
 & Forward velocity & 0.1\\
 & Turning velocity & 0.1\\
\hline
\multirow{4}{*}{Safety and Realism}
 & Root height & 0.05\\
 & Upper-body & 0.1\\
 & Nominal posture & 0.1\\
 & Joint velocities & 0.1\\
\hline
\end{tabular}
\end{center}
\end{table}



\section{Training for Challeging Terrains}

\begin{figure*}
  \includegraphics[width=\linewidth]{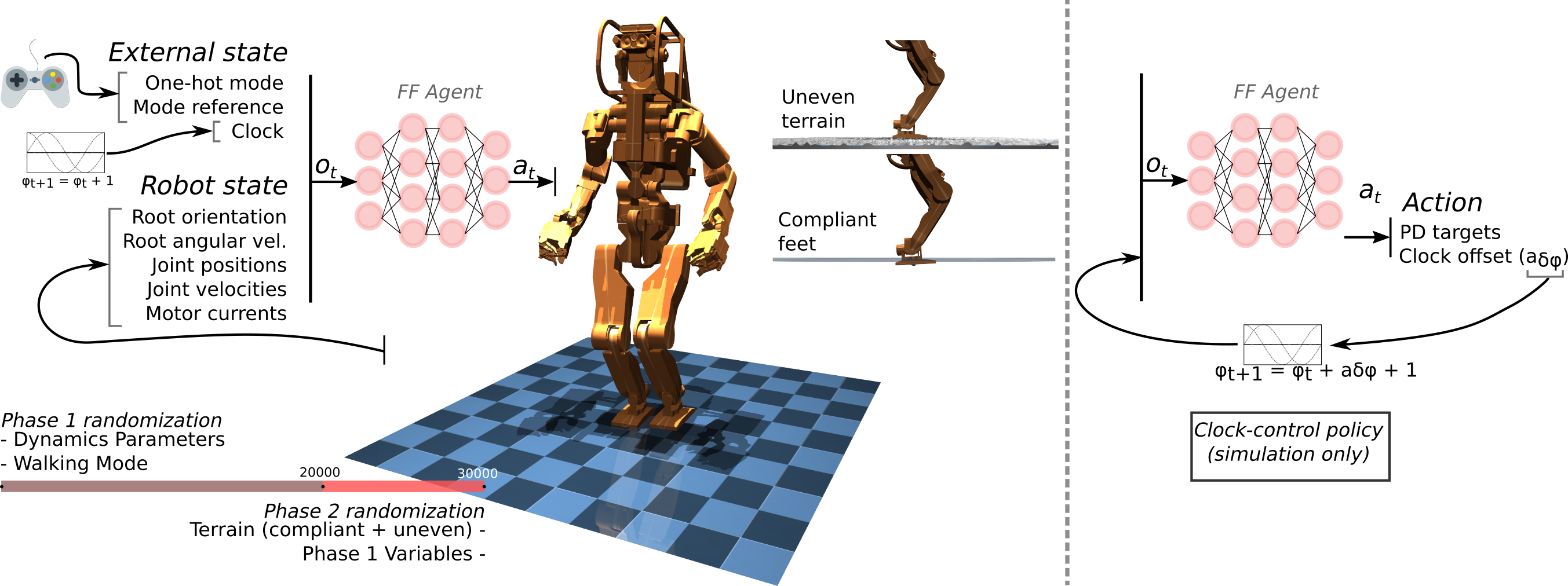}
  \caption{\textbf{Overview of our training framework}. (L) We propose to train
    a feedforward RL agent while exposing it to randomized dynamics parameters in the first
    phase and then, additionally, randomized uneven and compliant terrains in the second phase.
    The policy achieves zero-shot \textit{sim-to-real} transfer on the real HRP-5P.
    (R) We also propose an augmented policy that can make clock signal modifications for
    regulating the stepping frequency to achieve improved robustness on challenging terrains.}
  \label{figure:overview}
\end{figure*}


In this section, we first describe each of the key components of our pipeline for training
a robust policy in MuJoCo and for achieving real robot deployment. Then, in \autoref{sec:clock_control},
we introduce a new policy architecture that can predict clock signal modulations for aperiodic gait on
challenging terrain.

\subsection{Simulating Compliance}
We investigate how to simulate the foot sinking effect of walking on soft terrains in
the MuJoCo simulation environment by exploiting the physics engine's soft contact model.
MuJoCo allows users to set the solver parameters for each constraint independently to
achieve the desired behavior \cite{todorov2012mujoco}.

To this end, we adjust the \textit{time constant} parameter of the mass-spring-damper
model of the contact constraint between the feet and the ground bodies. We vary this
parameter in the range of $(0.02, 0.4)$ to simulate from a completely stiff to a more spring
like behavior of the feet contact with the ground. Notably, we found that it is more
convenient to set the \textit{time constant} parameter for the robot's feet ``geoms''
instead of the floor ``geoms'', especially because we also modify the ground profile
continuously during an episode to simulate an uneven terrain.


\subsection{Simulating Unevenness}
A commonly used approach for learning robust legged locomotion is to train the policy
in a simulation environment containing random discrete obstacles, slopes, and stairs.
Here the agent acts blindly and effectively learns recovery strategies by making continual
collisions with the environment, relying only on the proprioceptive state \cite{
  2020-science-blindQuadruped, siekmann2021blind}. However, as mentioned previously,
blind locomotion in the presence of large obstacles may be infeasible for HRP-5P.

In this work, we limit the peak height of obstacles on the floor to $\SI{4}{\cm}$ during training. We
simulate floor unevenness using the height fields functionality of MuJoCo.
At the start of the training we generate one height field of size $10m\times10m$ and a grid
size of $\SI{4}{\cm}$. This height
field is placed along with the default rigid, flat floor. Then, during the training
episodes, we randomize the $z$ position of this height field relative to the flat
floor. This creates the effect of generating a new terrain at each randomization with some
obstacles of varying height scattered on a flat floor at irregular intervals. The z-position
lies in the range of $(-\SI{4}{\cm}, \SI{0}{\cm})$, creating a completely flat floor at $z = -\SI{4}{\cm}$ and a fully
uneven terrain (i.e. no flat regions) at $z = \SI{0}{\cm}$. By pre-generating the height field in advance and randomizing its
position, we eliminate the need to generate a new height field at each episode, saving valuable
simulation time.

One limitation of this approach is that, during randomization of the terrain, the
uneven floor may move under the support foot in such a way so as to produce an upward
thrust. This is unlikely to happen in the real world. We mitigate this (partially)
by disabling terrain unevenness randomization in double-support phase, that is,
when both feet are on the ground.

\subsection{Curriculum Learning}
Our training procedure is shown in \autoref{figure:overview}L. The training is done
iteratively in two phases to achieve the final policy deployed on the real robot. First,
we train a base policy for all walking modes on a flat, rigid floor. The walking modes
are randomly switched between \textit{Standing}, \textit{Inplace}, \textit{Forward} at
every $5$s in an episode. However, the switch into and out from \textit{Standing} mode
is allowed only when the clock signal is the double-support region. The mode reference
value is simultaneously sampled at each switch from a uniform range of
$(-0.5, 0.5)\SI{}{\radian\per\s}$ turning speed if the mode is \textit{Inplace} and a
$(0.1, 0.4)\SI{}{\meter\per\s}$ forward walking speed if the mode is \textit{Forward}.

In the second phase, we finetune the base policy for challenging terrain by initiating
a new training process starting from the pre-trained network weights. The compliance
of each foot (``solref'' parameter in MuJoCo) is randomized after every $0.5$s on
average in the uniform range $(0.02, 0.4)$ in an episode. The randomization range was
determined through trial-and-error and visually inspecting the contact behavior of
the simulated robot. The 3D position of the uneven terrain (simulated using ``height fields''
in MuJoCo) is randomized at every $5$s inverval on average and is disabled during
double-support. The $x,y-$ positions are sampled uniformly from the range $(-0.5, 0.5)$m,
and $z-$ position from $(-0.04, 0)$m. Walking mode randomization is continued as in the
previous phase.

\subsection{Dynamics Randomization}
We found that it is much more effective to randomize a small set of
dynamics parameters of the robot model \textit{during an episode} at some reasonable
time interval, rather than at the start of the episode (conventional approach).
This can be expained by the fact that our simulation environment is unable to model
some aspects of the real robot dynamics realistically. For example, in the case of
joint friction, natively MuJoCo only allows us to simulate a simple friction model
(dry and Coulomb friction), while the real joint friction displays a much more complex
behavior. Thus, randomizing the friction coefficients during an episode will help prevent
the agent from overfitting to one specific friction dynamics.

We sample a new set of dynamics parameters on an average at every $0.5$s, which is shorter
than the length of one swing phase. Details about the parameters and corresponding sampling
ranges are provided in \autoref{table:dynrand}.

\section{Aperiodic Walking}
\label{sec:clock_control}

While the above mentioned modifications to the training environment are necessary
for training a policy for walking on challenging terrains, a reward function
based on a cyclic clock signal of constant period (\autoref{table:rewards})
limits the robot to a rigid stepping pattern. We argue that as a consequence of
this fixed, periodic stepping pattern, the robot is unable to overcome terrain
challenges that require large deviations from the prescribed swing and stance
durations. Such contact-timing deviations break the periodicity of bipedal
gait encoded in some form in most model-based and model-free approaches.
In other words, a fixed, periodic gait pattern acts as a hindrance to achieving
higher robustness.

We explore whether the control policy can be trained to achieve aperiodic walking
gait patterns and whether these aperiodic gaits can lead to higher robustness on uneven
terrain. To this end, we introduce a new policy that can predict clock signal modulations
such that new gait patterns may \textit{emerge} to adapt to different terrains and
command velocities (see \autoref{figure:overview}R).

Besides controlling the actuated joints of the robot we augment the policy output to control
the clock signal by predicting scalar offsets to the phase variable. Consequently, for this
policy the action space $\text{A} \in \mathbb{R}^{13}$, and the phase variable is computed as,

\begin{align}
\label{eq:phase_modulate}
  \phi_{t+1} = \phi_{t} + \text{clip}(a_{\delta \phi}, -5, 5) + 1
\end{align}

where, the phase offset action $a_{\delta \phi}$ is clipped to a conservative range $(-5, 5)$
timesteps. This range corresponds to a per-step maximum allowable offset of $5 \times 0.025$s = $0.125$s
at $40\SI{}{\hertz}$ control frequency.

This has been proposed previously for clock-based policies on bipeds, but it was found to have
negligible impact on the robot's performance \cite{siekmann2021blind}. However, we believe
that such clock-control offers a clear advantage for walking over uneven terrains and provide
simulation evaluations in \autoref{subsec:clock_control_effect} to study its impact.

\begin{table}[t]
\caption{Dynamics Randomization.}
\label{table:dynrand}
\begin{center}
\begin{tabular}{l l c}
\hline
 \multicolumn{1}{l}{\textbf{Parameter}} & \multicolumn{1}{l}{\textbf{Unit}} & \multicolumn{1}{l}{\textbf{Range}}\\
\hline
Joint damping coefficient & - & $(0.2, 5)$ \\
Joint static friction  & $\SI{}{\N-m}$ & $(2, 8)$ \\
Link mass & - & $[0.95, 1.05] \times$ default \\
Link CoM ($x$, $y$, $z$) & $\SI{}{\meter}$ & $\pm$0.01 from default \\
\hline
\end{tabular}
\end{center}
\end{table}

\section{Experiments}

\subsection{Implementation Details}
\textbf{Training.}
Our training hyperparameters are the same as \cite{singh2023learning}. Each episode rollout
spans a maximum of $\SI{10}{\second}$ of simulated time, and may reset if a terminal condition
is met. Training for 30000 iterations takes around 20 hours (on a AMD Ryzen Threadripper PRO 5975WX
CPU with 32 cores and 64 threads) to collect a total of 380 million samples for learning the full
task. Both the actor and critic policies are represented by MLP architectures to parameterize the
policy and the value function in PPO \cite{schulman2017proximal}. Both MLP networks have 2 hidden
layers of size 256 each and use \textit{ReLU} activations.

\textbf{Inference.}
The policy is wrapped in a C++ \textit{mc-rtc}\footnote{\url{https://jrl-umi3218.github.io/mc_rtc/index.html}}
controller and is evaluated in \textit{mc-mujoco} \cite{singh2023mc} before deployment. Controller
execution is on HRP-5P's onboard control PC (Intel NUC5i7RYH i7-5557U CPU with 2 cores, Ubuntu 18.04 LTS
PREEMPT-RT kernel). The inference is done at $\SI{40}{\hertz}$ with the PD controller running at
$\SI{1000}{\hertz}$.

The measured current feedback from the motors is fed to the control policy's observation after
transformation to the joint-level torque space. We refer the reader to \cite{singh2023learning}
for more details about current feedback for improved \textit{sim-to-real} transfer.

\subsection{Locomotion on Indoor and Outdoor terrains}
We created several test terrains inside the lab consisting of rigid irregular blocks, a
soft gym mattress, and a cushion foam block. From the 9 trials performed on the test
terrains, the proposed policy could succeed to traverse in 6 trials, that is, a success
rate of $67\%$. It is important to note the reason for the 3 failed cases. In the first
case, the policy triggered a conservative safety implemented in the robot middleware due
to excessive ankle motion. The safety limits were consequently relaxed. The other two
failures occurred while standing on a slope or stepping over obstacles higher than $\SI{4}{cm}$.

We noticed that the policy is prone to failure typically when in
standing mode (i.e. a very long double-support phase) with both feet on an inclined slope.
We suspect that this happens because the change in robot state due to incline occurs
gradually over time, and it becomes difficult for the policy to observe and compensate
for this slow change. Note that the policy was not trained for such a scenario, hence, we believe
terrain slope randomization at training time may resolve this issue. Similarly, the robot
struggled to step over obstacles of height greater than $\SI{4}{\cm}$, which is the maximum
height of unevenness during policy training. Training for obstacle heights beyond $\SI{4}{\cm}$
in simulation should be straightforward but may raise risks of robot damage during
\textit{sim-to-real}.

In the outdoor tests, the robot could successfully traverse a distance of approximately
$\SI{25}{\meter}$ on a paved street, and $\SI{30}{\meter}$ on an irregular grass lawn
area. The lawn provided a combination of irregularity and compliance due to small mounds and
depressions that was different from the test terrains available in the lab.

\textbf{Ablations.} We studied the importance of randomizing the terrain properties
during training using simulated and real robot experiments by evaluating four policies:
\begin{enumerate}
\item[(a)] \textit{Baseline}: trained only on flat floor
\item[(b)] \textit{Uneven terrain}: baseline policy finetuned on uneven but rigid terrain
\item[(c)] \textit{Fixed compliance}: baseline policy finetuned on uneven terrain with fixed compliance
(\textit{solref} parameter is set to 0.4)
\item[(d)] \textit{Terrain-randomized}: baseline policy finetuned on uneven terrain with randomized compliance
(\textit{solref} is in range $[0.02, 0.4]$).
\end{enumerate}

Only the \textit{baseline policy} is trained from scratch, for a total of 20000 training iterations (or about
250 millions samples). All the other policies are finetuned starting from the baseline for 10000
iterations.

Not surprisingly, the \textit{baseline policy} trained for a flat floor was unable to cope with
even a small height of unevenness in the simulation with a very short mean time before episode
termination --- $\SI{1.35}{\s}$ for $\SI{2}{\cm}$ and $\SI{5.6}{\s}$ for $\SI{1}{\cm}$ peak unevenness,
averaged over 20 episodes in each case. We did not attempt to walk over obstacles or soft terrain on the real robot
with this policy.

\textit{Uneven terrain} policy could succeed to walk on small obstacles ($\SI{3.5}{\cm}$) on
the simulated and real robot in several trials. However, since it was only trained on a rigid floor,
the robot failed to walk more than a few steps on the compliant terrain (soft blue mat) in $5/5$ trials.

\textit{Fixed compliance} policy could succeed to walk on $\SI{3.5}{\cm}$ rigid obstacles and
also on the compliant terrain. However, since this policy was trained without randomizing the
``softness'' of the floor, the real robot seemed to struggle to walk on the flat floor, and appeared
to make redundant (and dangerous) ankle twist motions.
We only performed one trial with this policy on the robot.

We achieved the best performance with the \textit{Terrain-randomized} policy. The policy appeared
to adapt the ankle motion depending on the compliance of the terrain while placing the foot horizontally
on the flat floor. We performed indoor and outdoor tests using this policy, with success rate noted above.

\subsection{Effect of Predicting Clock}
\label{subsec:clock_control_effect}
As described in \autoref{sec:clock_control}, we propose to augment the control policy's
action space such that it is able predict delta modifications to the observed clock signal.
We evaluate the behavior of the clock-control policy ($\text{A} \in \mathbb{R}^{13}$) and
compare it to the default policy ($\text{A} \in \mathbb{R}^{12}$) that cannot control the
clock. Both policies were trained for challenging terrain using the same routine of pre-training
on regular train followed by finetuning on randomized compliant and uneven terrain.

\begin{figure}[t]
  \begin{subfigure}{\linewidth}
    \includegraphics[width=\linewidth]{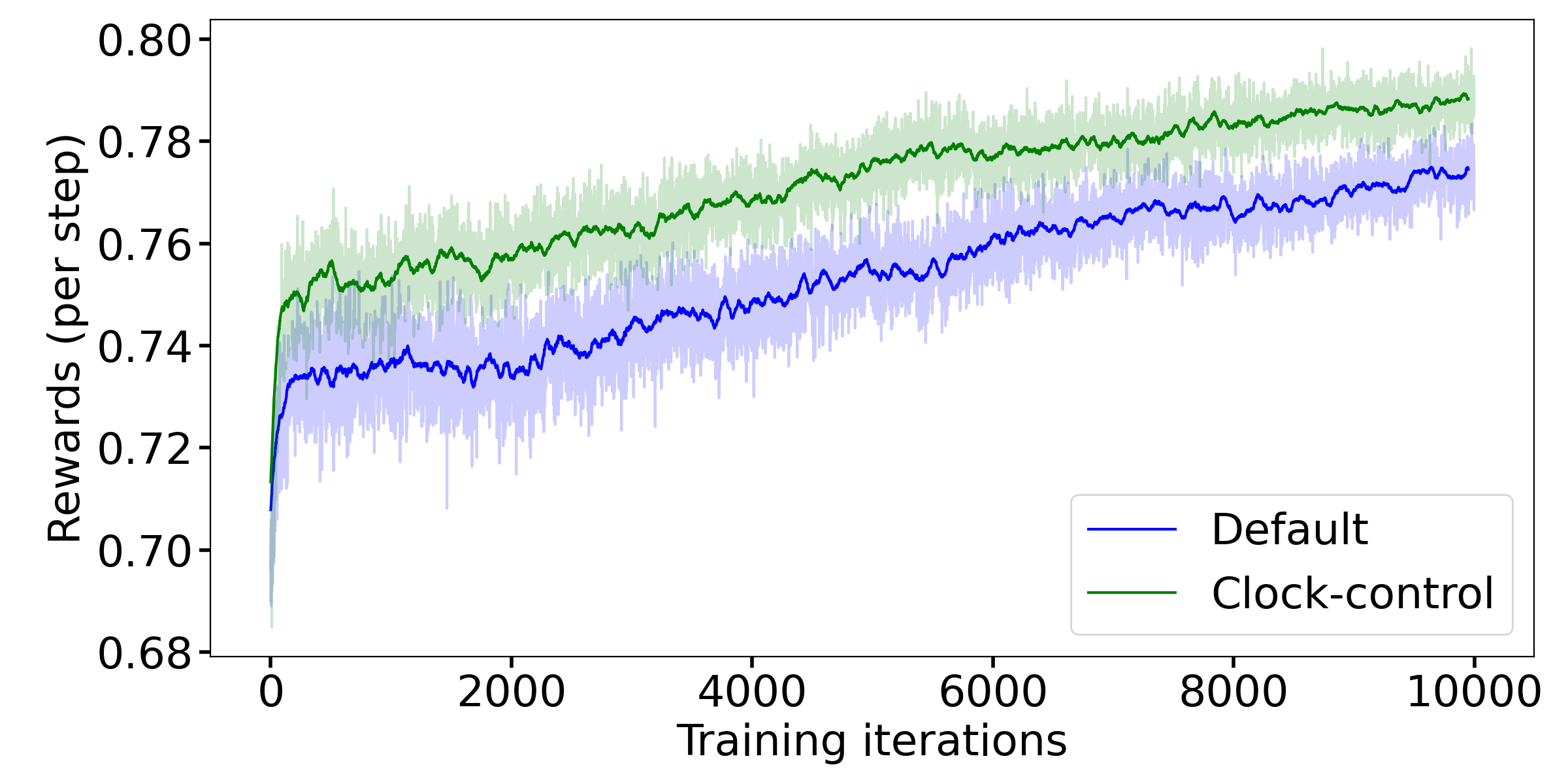}
    \caption{}
    \label{figure:clock_control_reward}
  \end{subfigure}
  \begin{subfigure}{\linewidth}
    \includegraphics[width=\linewidth]{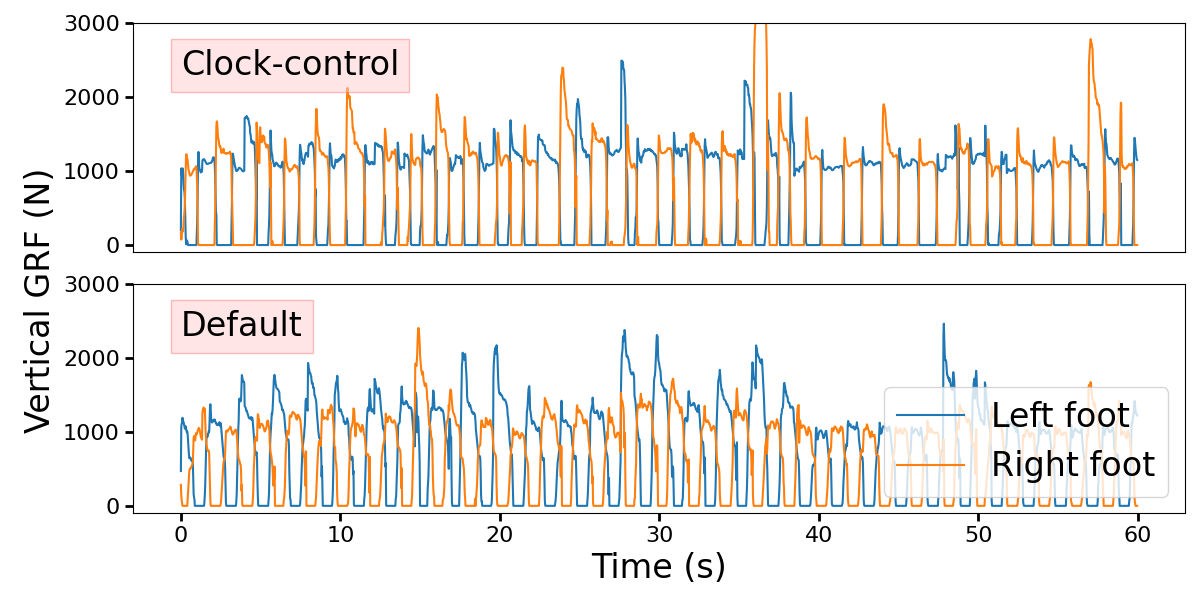}
    \caption{}
    \label{figure:clock_control_grf}
  \end{subfigure}
  \caption{\textbf{Behavior analysis of Clock Control (in simulation).}
    (a) Training reward curves for the clock-control policy and the default policy
    averaged over 3 training sessions with separate random seeds.
    Both policies are trained on randomized terrains starting from a regular terrain
    pre-trained policy. The clock-control policy converges to a higher reward.
    (b) 1-minute long episodes on the same uneven and soft terrain. The clock-control
    policy modifies the swing and stance duration in response to large disturbances while
    the default policy maintains a fixed gait pattern. \textbf{Note.} The sharp peak force
    for clock-control is due to inaccuracy in simulation (contact penetration). }
\end{figure}

First, we observed that the clock-control policy achieves a higher reward during training
compared to the default policy. \autoref{figure:clock_control_reward}) shows the training
reward curves averaged over 3 experiments with different random seeds. This is in alignment
to our expectations --- the ability to adapt footstep timing under the presence of external disturbances
provides greater flexibility from a control standpoint. This is better understood by analyzing the
ground reaction force (GRF) profiles of the two policies (\autoref{figure:clock_control_grf}). We
deployed a 60-second episode for stepping in-place with each policy and recorded the vertical GRF of
both feet on the same randomized terrain in simulation. The clock-control policy appears to modulate
the swing and support durations (regions of zero and non-zero force, respectively) of the gait while
the default policy exhibits a nearly regular gait pattern.

\begin{figure}[t]
  \includegraphics[width=\linewidth, height=4cm]{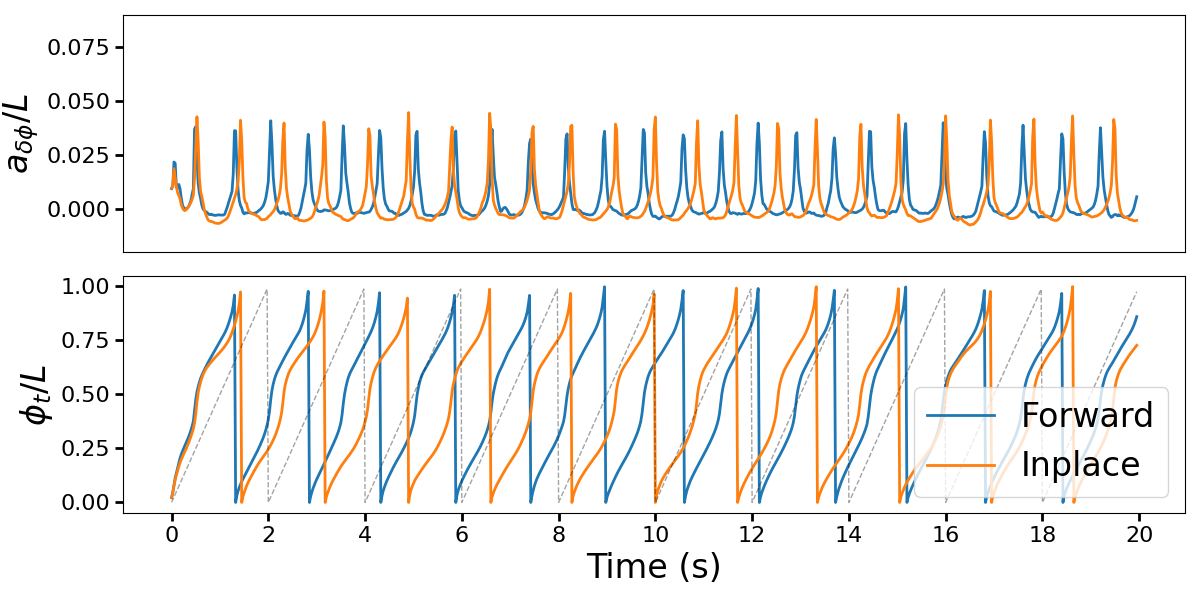}
  \caption{\textbf{Evolution of the phase variable} for a 20-second episode rollout on flat, rigid
    terrain for walking forward and stepping in-place with a clock-control policy.
    Policy phase offset predictions (top) and the actual scalar phase variable (bottom)
    are both normalized by a fixed period $L = 80$ for better visualization.
  }
  \label{figure:phase_mod_flat}
\end{figure}

Next, we record the value of the phase variable $\phi_{t}$ and the policy's phase offset
action $a_{\delta \phi}$ for simulated episodes on a regular, rigid terrain, for walking
forward at command speed of $0.4 \SI{}{\meter\per\s}$ and for stepping in-place (see
\autoref{figure:phase_mod_flat}). In either walking mode, we noticed that the policy learned
to speed up the clock (default gait cycle duration was $2$s). The average gait cycle duration
was $\SI{1.55}{\s}$ walking forwards at $0.4 \SI{}{\meter\per\s}$ (which includes accelerating
from start) and $\SI{1.71}{\s}$ while stepping in-place. This shows that the clock-control
policy prefers a shorter gait cycle duration in general, and an even shorter duration for walking
at higher speeds. Since a shorter gait cycle implies more frequent stepping, the policy's behavior
could be attributed to its increased ability react faster to disturbances. We also noticed
that policy mainly makes positive corrections to the clock even though it is allowed to
rewind the time.

The advantage of the clock-control policy specifically for locomotion over challenging terrain can
also be observed in terms of success rate on different terrains. \autoref{table:ep_lengths} lists
the mean time before termination for both policies at increasing peak height of unevenness with randomized foot
compliance. The mean values are computed over 100 simulation episodes in each case (800 episodes in
total) with each episode having a maximum length of $\SI{10}{\s}$. Both policies were trained for
a maximum unevenness height of $\SI{4}{\cm}$, and consequently, have similarly high success on
$\SI{4}{\cm}$ terrains. However, the clock-control policy outperforms the default policy when tested
on terrains with higher unevenness.

\textbf{Limitations.} We observed that real robot deployment of the clock-control policy is significantly
more difficult than the default policy (fixed clock). As this modification increases the unpredictability
of the robot's motion that requires implementation of additional safety measures, we skip the
evaluations of the clock-control policy on the real hardware in this work. However, our preliminary
investigations show that the \textit{sim-to-real} gap may be higher for the clock-control policy
than for the default policy. This could be due to poor system identification of the robot joints, leading to
the policy's inability to make fine adjustments to contact timings and durations on the real robot.

\begin{table}[t]
\caption{Mean Episode Length on Challenging Terrain\\ (10s max length)}
\label{table:ep_lengths}
\begin{center}
\begin{tabular}{c c c}
\hline
\multicolumn{1}{l}{\textbf{Unevenness height}($\SI{}{cm}$)} &
\multicolumn{1}{l}{\textbf{Default}($\SI{}{\s}$)} &
\multicolumn{1}{l}{\textbf{Clock-Control}($\SI{}{\s}$)}\\
\hline
$4$ & $9.925$ & $9.925$ \\
$5$ & $9.675$ & $\mathbf{9.725}$ \\
$6$ & $8.875$ & $\mathbf{9.525}$ \\
$7$ & $6.975$ & $\mathbf{8.4875}$ \\
\hline
\end{tabular}
\end{center}
\end{table}

\section{Conclusion}

In this work, we explored the application of model-free deep reinforcement learning for
developing humanoid locomotion controllers, specifically, for walking on challenging
terrains.

We showed that a relatively simple policy architecture and learning framework can attain
promising locomotion performance on the real robot. In a simulation environment, we propose
to first train a base policy on flat and rigid terrain, and then finetune the base policy
on a uniformly randomized curriculum of compliant and uneven terrains. The RL agent comprises
of a feedforward MLP network without incorporating observation history.

While it is difficult to quantitatively evaluate the wide range of existing humanoid locomotion
controllers, in our experiments we found the proposed RL framework to have achieved remarkably
robust locomotion performance. We were able to perform 2 outdoor experiments
--- without any falls in either trial or any other notable performance issues. Indoor
locomotion over the soft cushion object also displayed a robust success rate. Our real robot
experiments also help validate the use of current feedback for achieving \textit{sim-to-real}
transfer \cite{singh2023learning} on the HRP-5P robot.

We also proposed a new method that can allow the policy to make modification to the observed
clock signal. Through simulation experiments, we analyzed the behavior of this policy
on different terrains and command velocities and observed the \textit{emergence} of aperiodic
gait patterns leading to improved locomotion robustness.


We release the source code on GitHub for simulating compliant and uneven terrain and for
achieving omnidirectional walking behavior (with our existing RL training and evaluation
framework) for better reproducibility of our work.

\textbf{Future work.}
As the next step, we plan to deploy the clock-control policy on the real robot after introducing relevant
safety measures and better system identification of the joint dynamics. We also hope to identify and overcome
remaining factors for improving \textit{sim-to-real} transfer such as link compliance and state estimation
errors.


\section*{Acknowledgements}
The authors thank all members of JRL for providing their support in conducting robot
experiments that were done during the production of this work. This work was partially
supported by JST SPRING Fellowship Program, Grant Number JPMJSP2124 and JSPS KAKENHI
Grant Number JP22H05002.

\balance
\bibliographystyle{IEEEtran}
\bibliography{IEEEabrv,bibliography.bib}
\end{document}